%% file: continuous-domain-convolutions.tex
\documentclass[runningheads]{llncs}
\usepackage{graphicx}
\usepackage{amsmath,amssymb} 
\usepackage[svgnames]{xcolor}
\usepackage{color}

\usepackage{makecell}
\usepackage{multirow}
\usepackage{svg}
\usepackage{subcaption}
\usepackage{booktabs}
\usepackage{tikz}
\usepackage{lipsum}

\usepackage[ruled,vlined]{algorithm2e}

\usetikzlibrary{decorations.markings}
\usetikzlibrary{shapes.geometric}

\pgfdeclarelayer{edgelayer}
\pgfdeclarelayer{nodelayer}
\pgfsetlayers{edgelayer,nodelayer,main}

\tikzstyle{none}=[inner sep=0pt]

\definecolor{light-gray}{gray}{0.95}

\tikzstyle{rn}=[circle,fill=Red,draw=Black]
\tikzstyle{gn}=[circle,fill=Lime,draw=Black]
\tikzstyle{yn}=[circle,fill=Yellow,draw=Black]
\tikzstyle{bn}=[circle,fill=black!20,draw=Black]

\def\ACCV20SubNumber{904}  

\renewcommand{\xi}{x_i^\prime}
\newcommand{\Fii}{F^\prime}
\newcommand{\x}{x}

\newcommand{\xj}{x_j}
\newcommand{\pmm}{p_m}
\newcommand{\thetam}{\theta_m}
\newcommand{\Thetam}{\Theta^{(m)}}
\newcommand{\fii}{f_i^\prime}

\newcommand{\Fjj}{F}
\newcommand{\fjj}{f_j}
\newcommand{\dij}{\Delta x}

\renewcommand{\Xi}{\mathcal{X}^\prime}
\newcommand{\Xj}{\mathcal{X}}
\newcommand{\Ni}{N^\prime}

\newcommand{\Nm}{N^{(m)}}
\newcommand{\knn}{$k$NN}
\newcommand{\Si}{S^\prime}
\newcommand{\Sj}{S}
\newcommand{\ie}{\textit{i.e.}}
\newcommand{\eg}{\textit{e.g.}}
\newcommand{\etc}{\textit{etc.}}

\newcommand{\etal}{\textit{et al.}} 

\newcommand\blfootnote[1]{%
  \begingroup
  \renewcommand\thefootnote{}\footnote{#1}%
  \addtocounter{footnote}{-1}%
  \endgroup
}

\begin{document}
\pagestyle{headings}
\mainmatter

\title{Sparse Convolutions on Continuous Domains for \\Point Cloud and Event Stream Networks}




\titlerunning{Sparse Convolutions on Continuous Domains}
%
\author{Dominic Jack\inst{1}\orcidID{0000-0002-8371-3502} \and
Frederic Maire\inst{1}\orcidID{0000-0002-6212-7651} \and
\quad Simon Denman\inst{1}\orcidID{0000-0002-0983-5480} \and
Anders Eriksson\inst{2}\orcidID{0000-0003-2652-7110}}
%
\authorrunning{D. Jack et al.}
%
\institute{Queensland University of Technology, QLD, Australia\\
\email{thedomjack@gmail.com,\{f.maire,s.denman\}@qut.edu.au}\and
University of Queensland, QLD, Australia\\
\email{a.eriksson@uq.edu.au}}

\maketitle

\begin{abstract}
Image convolutions have been a cornerstone of a great number of deep learning advances in computer vision. The research community is yet to settle on an equivalent operator for sparse, unstructured continuous data like point clouds and event streams however. We present an elegant sparse matrix-based interpretation of the convolution operator for these cases, which is consistent with the mathematical definition of convolution and efficient during training. On benchmark point cloud classification problems we demonstrate networks built with these operations can train an order of magnitude or more faster than top existing methods, whilst maintaining comparable accuracy and requiring a tiny fraction of the memory. We also apply our operator to event stream processing, achieving state-of-the-art results on multiple tasks with streams of hundreds of thousands of events.
\end{abstract}

\section{Introduction}
\blfootnote{This research was supported by the Australian Research Council through the grant ARC FT170100072.}
Deep learning has exploded in popularity since AlexNet \cite{krizhevsky2012imagenet} achieved ground-breaking results in image classification \cite{deng2009imagenet}. The field now boasts state-of-the-art performance in fields as diverse as medical imaging \cite{erickson2017machine}, natural language processing \cite{devlin2018bert}, and molecular design \cite{elton2019deep}.

Robotics \cite{pierson2017deep} applications are of particular interest due to their capacity to revolutionize society in the near future. Driverless cars \cite{grigorescu2019survey} specifically have attracted enormous amounts of research funding, with advanced systems being built with multi-camera setups \cite{heng2019project}, active LiDAR sensors \cite{li2016vehicle}, and sensor fusion approaches \cite{gao2018object}.

At the other end of the spectrum, small mobile robotics applications and mobile devices benefit from an accurate 3D understanding of the world. These platforms generally don't have access to large battery stores or computationally hefty hardware, so efficient computation is essential. Even where compute is available, the cost of energy alone can be prohibitive, and the research community is beginning to appreciate the environmental cost of training massive power-hungry algorithms in data centres \cite{garcia2019estimation}.

The convolution operator has been a critical component of almost all recent advances in deep learning for computer vision. However, implementations designed for use with images cannot be used for data types that are not defined on a regular grid. Consider for example event cameras, a new type of sensor which shows great promise, particularly in the area of mobile robotics. Rather than reporting average intensities of every pixel at a uniform frame rate, pixels in an event camera fire individual events when they observe an intensity change. The result is a sparse signal with very fast response time, high dynamic range and low power usage. Despite the potential, this vastly different data encoding means that a traditional 2D convolution operation is no longer appropriate.
\begin{figure}
    \centering
    \includegraphics[width=0.9\textwidth]{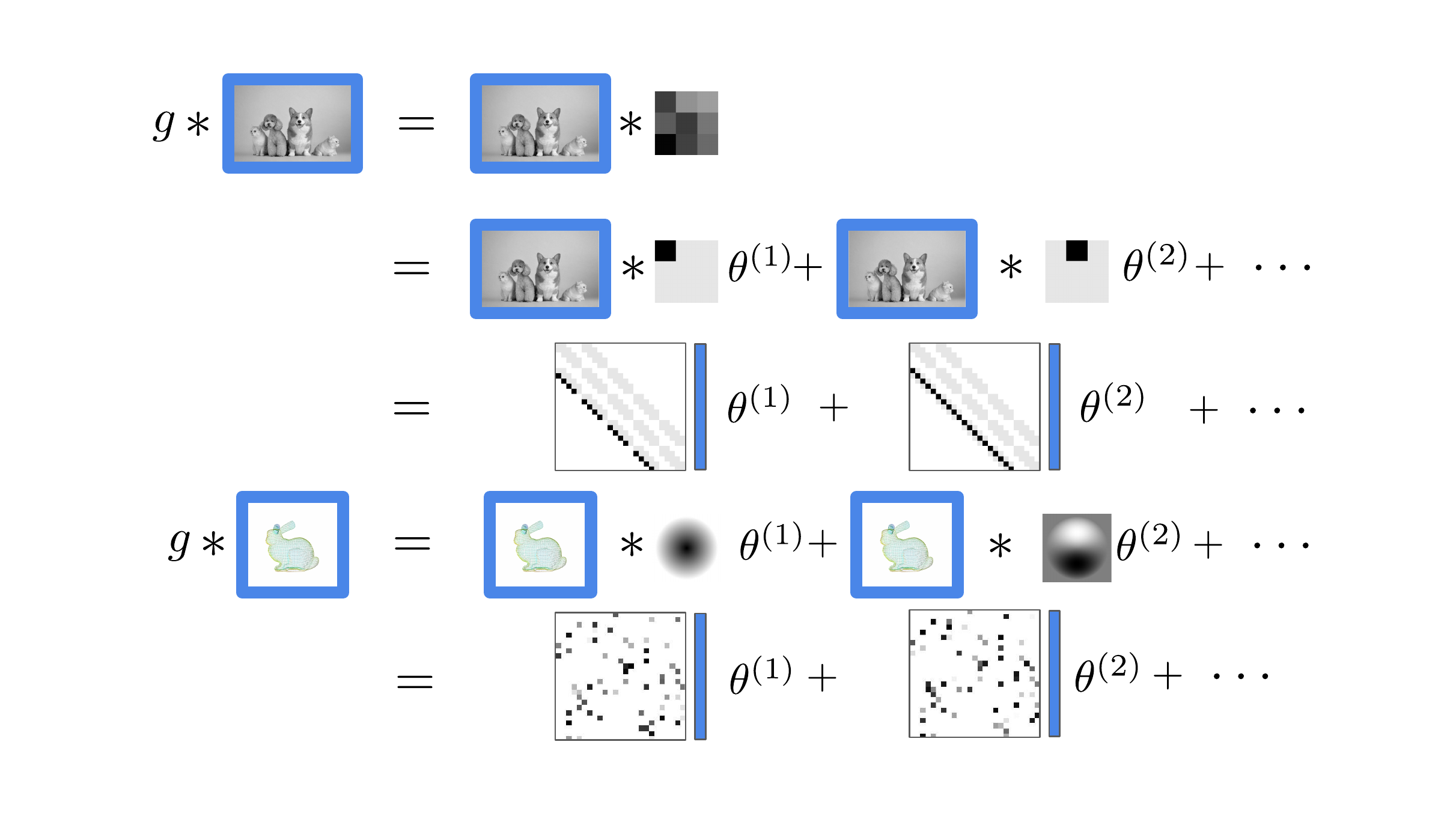}
    \caption{Learned image convolutions can be thought of as linear combinations of static basis convolutions, where the linear combination is learned. Each basis convolution can be expressed as a sparse-dense matrix product. We take the same approach with point clouds and event streams.}
    \label{fig:sparse-vis}
\end{figure}

In this work, we investigate how the convolution operator can be applied to two non-image input sources: point clouds and event streams. In particular, our contributions are as follows.
\begin{enumerate}
    \item We implement a convolution operator for sparse inputs on continuous domains using only matrix products and addition during training. While others have \textit{proposed} such an operator, we believe we are the first to \textit{implement} one without compromising the mathematical definition of convolution.
    \item We discuss implementation details essential to the feasible training and deployment of networks using our operator on modest hardware. We demonstrate speed-ups of an order of magnitude or more compared to similar methods with a memory foot-print that allows for batch sizes in the thousands.
    \item For point clouds, we discuss modifications that lead to desirable properties like robustness to varying density and continuity, and demonstrate that relatively small convolutional networks can perform competitively with much larger, more expensive networks.
    \item For event streams, we demonstrate that convolutions can be used to learn features from spiking network spike trains. By principled design of our kernels, we propose two implementations of the same networks: one for learning that takes advantage of modern accelerator hardware, and another for asynchronous deployment which can provide features or inferences associated with events as they arrive. We demonstrate the effectiveness of our learned implementation by achieving state-of-the-art results on multiple classification benchmarks, including a 44\% reduction in error rate on sign language classification \cite{bi2019graph}.
\end{enumerate}
\section{Prior Work}
\subsubsection{Point Clouds}
Early works in point cloud processing -- Pointnet \cite{qi2017pointnet} and Deep Sets \cite{zaheer2017deep} -- use point-wise shared subnetworks and order invariant pooling operations. The successor to Pointnet, Pointnet++ \cite{qi2017pointnet++} was (to the best of our knowledge) the first to take a hierarchical approach, applying Pointnet submodels to local neighborhoods.

SO-Net \cite{li2018so} takes a similar hierarchical approach to Pointnet++, though uses a different method for sampling and grouping based on self-organizing maps. DGCNN \cite{wang2019dynamic} applies graph convolutions to point clouds with edges based on spatial proximity. KCNet \cite{shen2018mining} uses dynamic kernel points in correlation layers that aim to learn features that encapsulate the relationships between those kernel points and the input cloud. While most approaches treat point clouds as unordered sets by using order-invariant operations, PointCNN \cite{li2018pointcnn} takes the approach of learning a canonical ordering over which an order-dependent operation is applied. SpiderCNN \cite{xu2018spider} and FlexConv \cite{groh2018flex} each bring their own unique interpretation to generalizing image convolutions to irregular grids. While SpiderCNN focuses on large networks for relatively small classification and segmentation problems, FlexConv utilizes a specialized GPU kernel to apply their method to point clouds with millions of points.



\subsubsection{Event Stream Networks}
Compared to standard images, relatively little research has been done with event networks. Interest has started to grow recently with the availability of a number of event-based cameras \cite{posch2010qvga,serrano2013128} and publicly available datasets \cite{serrano2013128,orchard2015converting,maqueda2018event,sironi2018hats,bi2019graph}.

A number of approaches utilize the extensive research in standard image processing by converting event streams to images \cite{maqueda2018event,nguyen2019real}. While these can leverage existing libraries and cheap hardware optimized for image processing, the necessity to accumulate synchronous frames prevents them from taking advantage of many potential benefits of the data format. Other approaches look to biologically-inspired spiking neural networks (SNNs) \cite{bohte2002error,russell2010optimization,cao2015spiking}. While promising, these networks are difficult to train due to the discrete nature of the spikes.

Other notable approaches include the work of Lagorce \etal\, \cite{lagorce2016hots}, who introduce hierarchical time-surfaces to describe spatio-temporal patterns; Sironi \etal\,\cite{sironi2018hats}, who show histograms of these time surfaces can be used for object classification; and Bi \etal \,\cite{bi2019graph}, who use graph convolution techniques operating over graphs formed by connecting close events in space-time.

\subsubsection{Sparse Convolutions}
Sparse convolutions have been used in a variety of ways in deep learning before. Liu \etal \cite{liu2015sparse} and Park \etal \cite{park2016faster} demonstrate improved speed from using implementations optimized for sparse kernel on discrete domains, while there are various voxel-based approaches \cite{graham2017submanifold,graham20183d,choy20194d} that look at convolutions on discrete sparse inputs and dense kernels. Other approaches involve performing dense discrete convolutions on interpolated projections \cite{jampani2016learning,su2018splatnet}.

\section{Method Overview}
For simplicity, we formulate continuous domain convolutions in the context of physical point clouds in Section \ref{sec:point-cloud-convolutions}, before modifying the approach for event streams in Section \ref{sec:event-stream-convolutions}. A summary of notation used in this section is provided in the supplementary material.

\subsection{Point Cloud Convolutions}\label{sec:point-cloud-convolutions}

We begin by considering the mathematical definition of a convolution of a function $h$ with a kernel $g$,
\begin{equation}
    (h * g)(t) = \int_{\mathcal{D}} h(\tau) g(t - \tau)\,d\tau.
\label{eqn:conv-def}
\end{equation}

We wish to evaluate the convolution of a function with values defined at fixed points $\xj$ in an input cloud $\Xj$ of size $\Sj$, at a finite set of points $\xi$ in an output cloud $\Xi$ of size $\Si$. We denote a single feature for each point in these clouds $f \in \mathbb{R}^{\Sj}$ and $f^\prime \in \mathbb{R}^{\Si}$ respectively. For the moment we assume coordinates for both input and output clouds are given. In practice it is often the case that only the input coordinates are given. We discuss choices of output clouds in subsequent sections.

By considering our convolved function $h$ to be the sum of scaled Dirac delta functions $\delta$ centred at the point coordinates,
\begin{equation}
    h(\x) = \sum_j \fjj \delta(\x - \xj),
\end{equation}
\noindent Equation \ref{eqn:conv-def} reduces to
\begin{equation}
    \fii = \sum_{\xj \in \mathcal{N}_i}\fjj g(\xi - \xj),
\label{eqn:conv-finite}
\end{equation}
\noindent where $\mathcal{N}_i$ is the set of points in the input cloud within some neighborhood of the output point $\xi$. We refer to pairs of points $\{\xj, \xi\}$ where $\xj \in \mathcal{N}_i$ as an \textit{edge}, and the difference in coordinates $\Delta x_{ij} = \xi - \xj$ as the \textit{edge vector}.

Like Groh \etal \cite{groh2018flex}, we use a kernel made up of a linear combination of $M$ unlearned basis functions $\pmm$,
\begin{equation}
    g(\dij; \theta) = \sum_m \pmm(\dij) \thetam,
\end{equation}
\noindent where $\thetam$ are learnable parameters. As with Groh \etal, we use geometric monomials for our basis function. Substituting this into Equation \ref{eqn:conv-finite} and reordering summations yields
\begin{equation}
    \fii = \sum_m \sum_{\xj \in \mathcal{N}_i} \pmm(\Delta x_{ij}) \fjj \thetam.
\label{eqn:conv-sparse-dense}
\end{equation}

We note the inner summation can be expressed as a sparse-dense matrix product,
\begin{equation}
    f^\prime = \sum_m \Nm f \theta_m,
\end{equation}
\noindent This is visualized in Figure \ref{fig:sparse-vis}. Neighborhood matrices $\Nm$ have the same sparsity structure for all $m$. Values $n_{ij}^{(m)}$ are given by the corresponding basis functions evaluated at edge vectors,
\begin{equation}
    n^{(m)}_{ij} = \begin{cases}
      \pmm(\Delta x_{ij}) & \xj \in \mathcal{N}_i, \\
      0 & \text{otherwise}.
   \end{cases}
\end{equation}

Generalizing to multi-channel input and output features $\Fjj \in \mathbb{R}^{\Sj \times Q}$ and $\Fii \in \mathbb{R}^{\Si \times P}$ respectively, this can be expressed as a sum of matrix products,
\begin{equation}
    \Fii = \sum_m \Nm \Fjj \Thetam,
\label{eqn:elegant}
\end{equation}
where $\Thetam \in \mathbb{R}^{Q \times P}$ is a matrix of learned parameters.

The elegance of this representation should not be understated. $\Nm$ is a sparse matrix defined purely by relative point coordinates and choice of basis functions. $\Thetam$ is a dense matrix of parameter weights much like traditional convolutional layers, and $\Fii$ and $\Fjj$ are feature matrices with the same structure, allowing networks to be constructed in much the same way as image CNNs.

We now identify three implementations with analogues to common image convolutions. A summary is provided in Table \ref{tab:implementation-summary}.

\paragraph{Down-Sampling Convolutions} Convolutions in which there are fewer output points than input points and more output channels than input channels are more efficiently computed left-to-right, \ie\, as $(\Nm F)\Theta^{(s)}$. These are analogous to conventional strided image convolutions.

\paragraph{Up-Sampling Convolutions} Convolutions in which there are more output points than input points and fewer output channels than input channels are more efficiently computed right-to-left, i.e. $\Nm \left(F \Thetam\right)$. These are analogous to conventional fractionally strided or transposed image convolutions.

\paragraph{Featureless Convolutions} The initial convolutions in image CNNs typically have large receptive fields and a large increase in the number of filters. For point clouds, there are often no input features at all -- just coordinates. In this instance the convolution reduces to a sum of kernel values over the neighborhood. In the multi-input/output channel context this is given by
\begin{equation}
	Z = \tilde{G} \Phi_0,
\end{equation}
\noindent where $\Phi_0 \in \mathbb{R}^{S \times Q}$ is the learned matrix and $\tilde{G} \in \mathbb{R}^{\Ni \times S}$ is a dense matrix of summed monomial values
\begin{equation}
	\tilde{g}_{is} = \sum_j \hat{n}_{ij}^{(m)}.
\end{equation}
\begin{table}[ht!]
    \vspace{-10px}
    \centering
    \begin{tabular}{l|c|c|c|c}
        \toprule
             & Opt. Cond. & Form & Mult. Adds & Mem. \\
        \hline
        In Place & \makecell{$Q = P$ \\ $\Si = \Sj$} & $\sum_m \Nm \Fjj \Thetam$  & $MP(E + SP)$ &  $SP$  \\
        \hline
        Down-Sample & \makecell{$Q < P$ \\ $\Si < \Sj$} & $\sum_m \left(\Nm F\right) \Thetam$  & $MQ(E + \Si P)$ & $\Si Q$ \\
        \hline
        Up-Sample & \makecell{$Q > P$ \\ $\Si > \Sj$} & $\sum_m \Nm \left(F \Thetam \right)$  & $MP(E + \Sj Q)$ & $\Sj P$ \\
        \hline
        Featureless & \makecell{$F = 1$} & $\tilde{G}\Phi_0$ & $M\Sj P$ & - \\
        \hline
    \end{tabular}
    \vspace*{1mm}
    \caption{Time complexity of different point cloud convolution operations and theoretical space complexity of intermediate terms (Mem). The matrix product for in place convolutions can be evaluated in either order.}
	\label{tab:implementation-summary}
\end{table}
\vspace{-30px}
\subsubsection{Neighborhoods}
To be consistent with the mathematical definition of convolution, the neighborhood of each point should be fixed, which precludes the use of $k$-nearest neighbors (\knn), despite its prevalence in the literature \cite{groh2018flex,xu2018spider,qi2017pointnet++,boulch2020convpoint}. The obvious choice of a neighborhood is a ball. Equation \ref{eqn:elegant} can be implemented trivially using either \knn\, or ball neighborhoods, though from a deep learning perspective each neighborhood has its own advantages and disadvantages.

\paragraph{Predictable computation time:} The sparse-dense matrix products have computation proportional to the number of edges. For \knn\, this is proportional to the output cloud size, but is less predictable when using ball-searches.

\paragraph{Robustness to point density:} Implementations based on each neighborhood react differently to variations in point density. As the density increases, \knn\, implementations shrink their receptive field. On the other hand, ball-search implementations suffer from increased computation time and output values proportional to the density.

\paragraph{Discontinuity in point coordinates:} Both neighborhood types result in operations that are discontinuous in point coordinates. \knn\, convolutions are discontinuous as the $k^{\text{th}}$ and $(k+1)^{\text{th}}$ neighbors of each point pass each other. Ball-search convolutions have a discontinuity at the ball-search radius.

\paragraph{Symmetry:} Connectedness in ball-neighborhoods is symmetric -- i.e. if $\xi \in \mathcal{N}_j$ then $\xj \in \mathcal{N}_i$ for neighborhood functions with the same radius. This means the neighborhood matrix $N_{IJ}$ between sets $\mathcal{X}_I$ and $\mathcal{X}_J$ is related to the reversed neighborhood by $N_{IJ} = N_{JI}^T$ (up to a possible difference in sign due to the monomial value). This allows for shared computation between different layers.

\paragraph{Transposed Neighborhood Occupancy:} For \knn, all neighborhood matrices are guaranteed to have $k$ entries in each row. This guarantees there will be no empty rows, and hence no empty neighborhoods. Ball search neighborhoods do not share this property, and there is no guarantee points will have any neighbors. This is important for transposed convolutions, where points may rely on neighbors from a lower resolution cloud to seed their features. 

\subsubsection{Subsampling}
Thus far we have remained silent as to how the $\Si$ output points making up $\Xi$ are chosen. In-place convolutions can be performed with the same input and output clouds, but to construct networks we would like to reduce the number of points as we increase the number of channels in a similar way to image CNNs. We adopt a similar approach to Pointnet++ \cite{qi2017pointnet++} in that we sample a set of points from the input cloud. Pointnet++ \cite{qi2017pointnet++} selects points based on the first $\Si$ points in iterative farthest point (IFP) ordering, whereby a fixed number of points are iteratively selected based on being farthest from the currently selected points. For each point selected, the distance to all other points has to be computed, resulting in an $\mathcal{O}(\Si\Sj)$ implementation.

To improve upon this, we begin by updating distances only to points within a ball neighborhood -- a neighborhood that may have already been computed for the previous convolution. By storing the minimum distances in a priority queue, this sampling process still gives relatively uniform coverage like the original IFP, but can be computed in $\mathcal{O}(\Si \bar{k})$, where $\bar{k}$ is the average number of neighbors of each point.


We also propose to terminate searching once this set of neighborless candidates has been exhausted, rather than iterating for a fixed number of steps. We refer to this as \textit{rejection sampling}. This results in point clouds of different sizes, but leads to a more consistent number of edges in subsequent neighborhood matrices. It also guarantees all points in the input cloud will have a neighbor in the output cloud. We provide pseudo-code for these algorithms and illustrations in the supplementary material.

\subsubsection{Weighted Convolutions}
To address both the discontinuity at the ball radius and the neighbor count variation inherent to using balls, we propose using a weighted average convolution by weighting neighboring values by some continuous function $w$ which decreases to zero at the ball radius,
\begin{equation}
	\hat{n}^{(m)}_{ij} = \frac{1}{W_i} w_{ij} n^{(m)}_{ij}
\end{equation}
\noindent where $w_{ij} = w(|\Delta x_{ij}|)$ and $W_i = \sum_j w_{ij}$. We use $w(x) = 1 - x / r$ for our experiments, where $r$ is the search radius.

\subsubsection{Comparison to Existing Methods}
We are not the first to propose hierarchical convolution-like operators for learning on point clouds. In this section we look at a number of other implementations and identify key differences.

Pointnet++ \cite{qi2017pointnet++} and SpiderCNN \cite{xu2018spider} each use feature kernels which are non-linear with respect to the learned parameters. This means these methods have a large memory usage which increases as they create edge features from point features, before reducing those edge features back to point features.

Pointnet++ claims to use a ball neighborhood -- and show results are improved using this over \knn. However, their implementation is based on a truncated \knn search with fixed $k$, meaning padding edges are created in sparse regions and meaningful edges are cropped out in dense regions. The cropping is partially offset by the use of max pooling over the neighborhood and IFP ordering, since the first $k$ neighbors found are relatively spread out over the neighborhood. As discussed however, IFP is $\mathcal{O}(\Sj\Si)$ in time, but removing this means results in the truncated ball search will no longer necessarily be evenly distributed. Also, the padding of sparse neighborhoods leads to an inefficient implementation, as edge features are computed despite never being used.

FlexConv \cite{groh2018flex} present a very similar derivation to our own. However, they implement Equation \ref{eqn:conv-sparse-dense} with a custom GPU kernel that only supports \knn.

On the whole, we are unable to find any existing learned approaches that perform true ball searches, nor make any attempt to deal with the discontinuity inherent to \knn. We accept models are capable of learning robustness to such discontinuities, but feel enforcing it at the design stage warrants consideration.

\subsubsection{Data Pipeline}
There are two aspects of the data processing that are critical to the efficient implementation of our point cloud convolution networks.

\paragraph{Neighborhood Preprocessing} The neighborhood matrices $N^{(m)}$ are functions of relative point coordinates and the choice of unlearned basis functions -- they do not depend on any learned parameters. This means they can be pre-computed, either online on CPUs as the previous batch utilizes available accelerators, or offline prior to training. In practice we only pre-compute the neighborhood indices and calculate the relative coordinates and basis functions on the accelerator. This additional computation on the accelerator(s) takes negligible time and reduces the amount of memory that needs to be stored, loaded and shipped.

\paragraph{Ragged Batching} During the batching process, the uneven number of points in each cloud for each example can be concatenated, rather than padded to a uniform size, and sparse matrices block diagonalized. For environments where fixed-sized inputs are required, additional padding can occur at the \textit{batch} level, rather than the individual example level, where variance in the average size will be smaller.

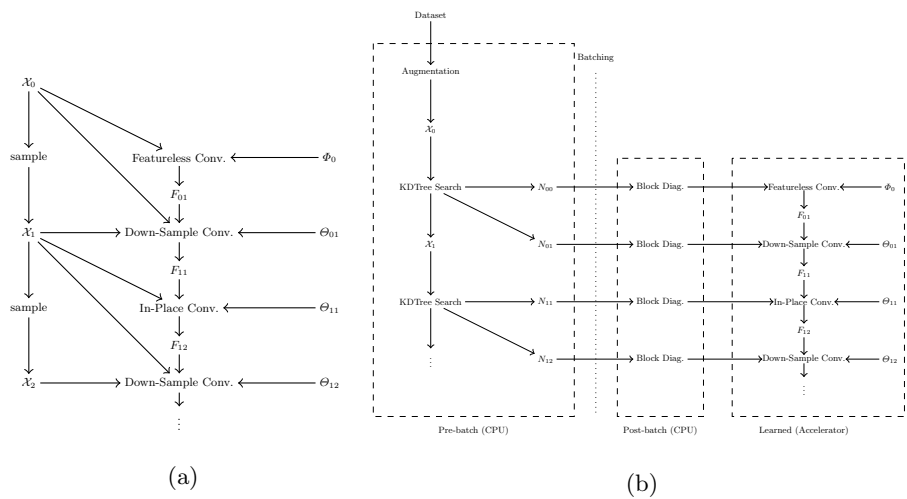
\begin{figure}[ht!]
    \centering
    \begin{subfigure}{.4\textwidth}
        \centering
        \vspace{22px}
        \input{tikz/conceptual.tikz}
        \vspace{5px}
        \caption{}
    \end{subfigure}
    \begin{subfigure}{.6\textwidth}
        \resizebox{\columnwidth}{!}{
        \input{tikz/meta.tikz}
        }
        \vspace{-5pt}
        \caption{}
    \end{subfigure}
    \caption{(a) Conceptual network vs (b) separate computation graphs.}
    \label{fig:meta-network}
\end{figure}

Unlike standard dataset preprocessing, our networks require network-specific preprocessing -- preprocessing dependent on \eg\, the size of the ball searches at each layer, the number of layers \etc \ To facilitate testing and rapid prototyping, we developed a meta-network module for creating separate pre- and post-batch preprocessing, while simultaneously building learned network stages based on a single conceptual network architecture. This is illustrated in Figure \ref{fig:meta-network}.
\subsection{Event Stream Convolutions}\label{sec:event-stream-convolutions}
Event streams from cameras can be thought of as 3D point clouds in $(x, y, t)$ space. However, only the most fundamental of physicists would consider space and time equivalent dimensions, and in practice their use cases are significantly different. For event cameras in particular,
\begin{itemize}
    \item spatial coordinates of events are discrete and occur on a fixed size grid corresponding to the pixels of the camera;
    \item the time coordinate is effectively continuous and unbounded; and
    \item events come in time-sorted order.
\end{itemize}

\noindent We aim to formulate a model with the following requirements:
\begin{itemize}
    \item \textit{Intermediate results}: we would like our model to provide a stream of predictions, updated as more information comes in, rather than having to wait for the end of a sequence before making an inference.
    \item \textit{Run indefinitely}: we would like to deploy these models in systems which can run for long periods of time. As such, our memory and computational requirements must be $\mathcal{O}(1)$ and $\mathcal{O}(E)$ respectively, with respect to the number of events.
\end{itemize}

Unfortunately, these requirements are difficult to enforce while making good use of modern deep learning frameworks and hardware accelerators. That said, just because we desire these properties in our end system does not mean we need them during training. By using convolutions with a domain of integration extending \textit{backwards in time only} and using an exponentially decaying kernel, we can train using accelerators on sparse matrices and have an alternative deployment implementation which respects the above requirements.

Formally, we propose neighborhoods defined over small spatial neighborhood of size $M_u$ -- similar to image convolutions -- extending backwards in time -- with a kernel given by
\begin{equation}
    g(u, \Delta t) = \sum_{v} \exp(-\lambda_{uv} \Delta t)\theta_{uv},
    \label{eqn:event-kernel}
\end{equation}
where $u$ corresponds to the pixel offset between events and $v$ sums over some fixed temporal kernel size $M_v$, $\lambda_{uv}$ is a learned temporal decay term enforced to be positive, $\theta_{uv}$ is a learned parameter and $u$ extends over the spatial neighborhood. A temporal domain of integration extending backwards in time only ensures $\Delta t \ge 0$, hence we ensure the effects of events on features decay over time.

\subsubsection{Dual Implementations}

For training, the kernel function of Equation \ref{eqn:event-kernel} can be used in Equation \ref{eqn:conv-sparse-dense} and reduced to a form similar to Equation \ref{eqn:elegant}, where $M = M_u M_v$. This can be implemented in the same way as our point cloud convolutions. Unfortunately, this requires us to construct the entire sparse matrix, removing any chance of getting intermediate results when they are relevant, and also breaks our $\mathcal{O}(1)$ memory constraint with respect to the number of events.

As such, we additionally propose a deployment implementation that updates features at pixels using exponential moving averages in response to events. As an input events come in, we decay the current values of the corresponding pixel by the time since it was last updated and add the new input features. When the features for an output event are required, the features of the pixels in the receptive field can be decayed and then transformed by $\Theta^{(uv)}$, and reduced like a standard image convolution. Formally, we initialize $\mathbf{z}^{(uv)}_x = \mathbf{0} \in \mathbb{R}^Q$ and $\tau_x=0$ for all pixels $x$. For each input event ($x, t)$ with features $\mathbf{f}$, we perform the following updates:
\begin{gather}
    \mathbf{z}^{(uv)}_x \leftarrow \mathbf{f} + \exp(-\lambda_{uv}(t - \tau_x)) \mathbf{z}^{(uv)}_x \\
    \tau_x \leftarrow t.
\end{gather}

Features $\mathbf{f}^\prime$ for output event at $(x^\prime, t^\prime)$ can thus be computed by
\begin{equation}
    \mathbf{f}^\prime = \sum_{u, v} \exp(-\lambda_{uv}(t^\prime - \tau_{x\prime-u})) {\mathbf{z}^{(uv)}_{x^\prime-u}}^T \Theta^{(uv)}.
\end{equation}

This requires $\mathcal{O}(M_u M_v Q)$ operations per input event, $\mathcal{O}(M_u M_v P Q)$ operations per output event and $\mathcal{O}(M_u M_v Q)$ space per pixel. Alternatively, the linear transform can be applied to $\mathbf{f}$ during the $\mathbf{z}^{(uv)}_x$ update (equivalent to up-sampling convolutions) for subtly different space and computational requirements. Either way, our requirements are satisfied.


\subsubsection{Subsampling}
As with our point cloud formulation, we would like a hierarchical model with convolutions joining multiple streams with successively lower resolution and higher dimensional features. We propose using an unlearned leaky-integrate-and-fire (LIF) model due to the simplicity of the implementation and its prevalence in SNN literature \cite{koch1998methods}.

LIF models transform input spike trains by tracking a theoretical voltage at each location or ``neuron". These voltages exponentially decay over time, but are increased discontinuously by input events in some receptive field. If the voltage at a location exceeds a certain threshold, the voltage at that neuron is reset, and an output event is fired. SNNs generally learn the sensitivity of each output neuron to input spikes. We take a simpler approach, using a fixed voltage increase of $1 / n$ as a result of an input spike, where $n$ is the number of output neurons affected by the input event. Note we do not suggest this is optimal for our use case -- particularly the unlearned nature of it -- though we leave additional investigation of this idea to future work.

\section{Experiments}
We perform experiments on various classification tasks across point clouds and event streams. We provide a brief overview of network structures here. Model diagrams and technical details about the training procedure are provided in the supplementary material.

We investigate our point cloud operator in the context of ModelNet40 \cite{wu20153d}, a 40-class classification problem with 9840 training examples and 2468 testing examples. We use the first 1024 points provided by Pointnet++ \cite{qi2017pointnet++} and use the same point dropout, random jittering and off-axis rotation, uniform scaling and shuffling data augmentation policies.

We construct two networks based loosely on Resnet \cite{he2016deep}. Our larger model consists of an in-place convolution with 32 output channels, followed by 3 alternating down-sampling and in-place residual blocks, with the number of filters increasing by a factor of 2 in each down-sampling block. Our in-place ball radii start at 0.1125 and increase by a factor of 2 each level. Our down-sample radii are $\sqrt{2}$ larger than the previous in-place convolution. This results in sampled point clouds with roughly 25\% of the original points on average, roughly 10 neighbors per in-place output point and 20 neighbors per down-sample output point. After our final in-place convolution block we use a single point-wise convolution to increase the number of filters by a factor of 4 before max pooling across the point dimension. We feed this into a single hidden layer classifier with 256 hidden units. All convolutions use monomial basis functions up to 2nd order. We use dropout, batch normalization and L2 regularization throughout. Our smaller model is similar, but skips the initial in-place convolution and has half the number of filters at each level. Both are trained using a batch size of 128 using Adam optimizer \cite{kingma2014adam} and with the learning rate reduced by a factor of 5 after 20 epochs without an improvement to training accuracy.

For event streams, we consider 5 classification tasks -- N-MNIST and N-Caltech101 from Orchard \etal \cite{orchard2015converting}, MNIST-DVS and CIFAR10-DVS from Serrano \etal \cite{serrano2013128}) and ASL-DVS from Bi \etal \cite{bi2019graph}.

All our event models share the same general structure, with an initial 3x3 convolution with stride 2 followed by alternating in-place resnet/inception-inspired convolution blocks and down-sample convolutions (also 3x3 with stride 2), finishing with a final in-place block. We doubled the number of filters and the LIF decay time at each down sampling.

The result is multiple streams, with each event in each stream having its own features. The features of any event in any stream could be used as inputs to a classifier, but in order to compare to other work we choose to pool our streams by averaging over $(x, y, t)$ voxels at our three lowest resolutions. For example, our CIFAR-10 model had streams with learned features at $64 \times 64$ down to $4 \times 4$. We voxelized the $16 \times 16$ stream to $16 \times 16 \times 4$, the $8 \times 8$ stream into an $8 \times 8 \times 2$ grid and the final stream into a $4 \times 4 \times 1$. Each voxel grid beyond the first receives inputs from the lower resolution voxel grid (via a strided $2 \times 2 \times 2$ voxel convolution), and from the average of the event stream. In this way, examples with relatively few events that result in zero events at the final stream still resulted in predictions (empty voxels are assigned the value $\textbf{0}$). Hyperparameters associated with stream propagation (decay rate, spike threshold and reset potential) were hand-selected via an extremely crude search that aimed to achieve a modest number of events in the final stream for most examples. These hyperparameters, along with further details on training and data augmentation are provided in the supplementary material.

\section{Results}
\subsection{Point Clouds}
We begin by benchmarking our implementations of Equation \ref{eqn:elegant}. We implement the outer summation in two ways: a parallel implementation which unstacks the relevant tensors and computes matrix-vector products in parallel, and a map-reduce variant which accumulates intermediate values. Both are written entirely in the high-level python interface to Tensorflow 2.0.

We compare with the work of Groh \etal \cite{groh2018flex} who provide benchmarks for their own tensorflow implementation, as well as a custom CUDA implementation that only supports \knn. Our implementations are written entirely in the high-level Tensorflow 2 python interface and can handle arbitrary neighborhoods. Computation time and memory requirements are shown in Table \ref{tab:flex-conv-benchmarks}. Values do not include neighborhood calculations. Despite our implementation being more flexible, our forward pass is almost an order of magnitude faster, and our full training pass is sped up more than 60-fold. Our implementation does require more memory. We also see significant improvements by using Tensorflow's accelerated linear algebra just-in-time (JIT) compilation module, particularly in terms of memory usage.
\begin{table}[ht!]
    \centering
    \begin{tabular}{lrrrr}
    \toprule
                  & \multicolumn{2}{c}{Time (ms)} & \multicolumn{2}{c}{GPU Mem (Mb)} \\
                  \hline
                  & Forward & Backward & Forward & Backward \\ 
                  \hline
         TF \cite{groh2018flex} & 1829 & 2738 & 34G & 63G \\
         Custom \cite{groh2018flex}           & 24.0 & 265.0  & \textbf{8.4} & \textbf{8.7} \\
         \hline
         \hline
         $(NF)\Theta$     & 2.9 & 5.1 & 57.3 & 105.0 \\
         $(NF)\Theta$-JIT & \textbf{2.7} & 5.0 & 41.0 & 41.0 \\
         $N(F\Theta)$     & 2.9 & 4.3 & 56.0 & 56.2 \\
         $N(F\Theta)$-JIT & \textbf{2.7} & \textbf{4.1} & 40.0 & 49.0 \\
    \bottomrule
    \end{tabular}
    \vspace*{1mm}
    \caption{Equation \ref{eqn:elegant} implementations vs. FlexConv benchmarks on an Nvidia GTX-1080Ti. $M = 4$, $P = Q = 64$, $\Sj = \Si = 4096$, 9 neighbors and batch size of $8$. Backward passes compute gradients w.r.t. learned parameters and input features ($F$ and $\Theta$). JIT rows correspond to just-in-time compiled implementations excluding compile time.}
    \label{tab:flex-conv-benchmarks}
\end{table}

Next we look at training times and capacity of our model on the ModelNet40 classification task using 1024 input points. Table \ref{tab:modelnet-benchmarks} shows performance at various possible batch sizes and training times for our standard model compared to various other methods. For fair comparison, we do not use XLA compilation.

\begin{table}[h!]
\centering
\makebox[0pt][c]{\parbox{1\textwidth}{%
    \begin{minipage}[t]{0.55\hsize}\centering
            \vspace{0pt}

            \centering
            \begin{tabular}{lr|rr}
            \toprule
                \multirow{2}{*}{Model} & \multirow{2}{*}{Batch Size} & \multicolumn{2}{l}{Epoch time (s)} \\
                \cmidrule(lr){3-4}
                &            & Online & Offline \\
                \hline
                SpiderCNN \cite{xu2018spider}      & 24 & 196 & -\\
                \hline
                \multirow{2}{*}{Pointnet++ \cite{qi2017pointnet++}} & 32 & 56 & - \\
                                                   & 64 & 56 & - \\
                \hline
                \multirow{3}{*}{PointCNN \cite{li2018pointcnn}} & 32 & 35 & - \\
                 & 64 & 33 & - \\
                 & 128 & 33 & - \\
                \hline
                \hline
                \multirow{5}{*}{Ours (large)}
                              &   32 & 12.9 & 6.80 \\
                              &   64 & 11.8 & 5.10 \\
                              &  128 & 11.2 & 4.13 \\
                              & 1024 & 12.4 & 3.60 \\
                              & 4096 & 11.5 & \textbf{3.35}  \\
                \hline
                \multirow{5}{*}{Ours (small)}          
                              &   32 & 12.0 & 4.35 \\
                              &   64 & 11.4 & 2.83 \\
                              &  128 & 11.2 & 2.06 \\
                              & 1024 & 11.4 & \textbf{1.38} \\
                              & 4096 & 11.5 & 1.41 \\
                              & \textbf{9840} & 13.0 & 1.39 \\
            \bottomrule
            \end{tabular}
            \vspace*{1mm}
            \caption{Time to train 1 epoch of ModelNet40 classification on an Nvidia GTX-1080Ti. Online/offline refers to preprocessing.}
            \label{tab:modelnet-benchmarks}

    \end{minipage}
    \hfill
    \begin{minipage}[t]{0.48\hsize}
        \vspace{0pt}
        \centering
            \begin{tabular}{l|r|r}
            \toprule
                Model     & Reported/Best & Mean  \\
                \hline
                \textbf{Ours (small)} & 88.77 & 87.94 \\
                Pointnet \cite{qi2017pointnet}  &             89.20 & 88.65 \\
                KCNet \cite{shen2018mining}    &             91.00 & 89.62 \\
                DeepSets \cite{zaheer2017deep} &             90.30 & 89.71 \\
                Pointnet++ \cite{qi2017pointnet++}&             90.70 & 90.14 \\
                \textbf{Ours (large)} & 91.08 & 90.34 \\
                DGCNN \cite{wang2019dynamic}    &             92.20 & 91.55 \\
                PointCNN \cite{li2018pointcnn} &             92.20 & 91.82 \\
                SO-Net \cite{li2018so}   &             \textbf{93.40} & \textbf{92.65} \\
            \bottomrule
            \end{tabular}
            \caption{Top-1 instance accuracy on ModelNet40, sorted by mean of 10 runs according to Koguciuk \etal \cite{koguciuk20193d}, for our large model with batch size 128. Reported/Best are those values reported by other papers, and the best of 10 runs for our models.}
            \label{tab:modelnet-accuracy}
    \end{minipage}
    \hfill
}}
\end{table}

Clearly our model runs significantly faster than those we compare to. Just as clear is the fact that our models which compute neighborhood information online are CPU-constrained. This preprocessing is not particularly slow -- a modest 8-core desktop is capable of completing the 7 neighborhood searches and 3 rejection samplings associated with each example on our large model at over 800 Hz, which results in training that is still an order of magnitude faster than the closest competitor -- but in the context of an accelerator-based training loop that runs at up to 3000 Hz this is a major bottleneck.

One might expect such a speed-up to come at the cost of inference accuracy. Top-1 accuracy is given in Table \ref{tab:modelnet-accuracy}. We observe a slight drop in performance compared to recent state-of-the-art methods, though our large model is still competitive with well established methods like Pointnet++. Our small model performs distinctly worse, though still respectably.

\subsection{Event Camera Streams}
Table \ref{tab:event-results} shows results for our method on the selected classification tasks. We see minor improvements over current state-of-the art methods on the straight-forward MNIST variants, though acknowledge the questionable value of such minor improvements on datasets like these. We see a modest improvement on CIFAR-10, though perform relatively poorly on N-Caltech101. Our ASL-DVS model significantly out-performs the current state-of-the-art, with a 44\% reduction in error rate. We attribute the greater success on this last dataset compared to others to the significantly larger number of examples available during training ($\sim$80,000 vs $\sim$10,000).

\begin{table}[ht!]
    \centering
    \begin{tabular}{lrrrrr}
    \toprule
    Model  & N-MNIST & MNIST-DVS & CIFAR-DVS & NCaltech101 & ASL-DVS \\
    \hline
    HATS \cite{sironi2018hats}   &  99.1  &  98.4 &     52.4 &      64.2  &  -      \\
    RG-CNN \cite{bi2019graph}    &  99.0  &  98.6 &     54.0 &      \textbf{65.7}  &   90.1 \\
    \hline
    Ours   &  \textbf{99.2}  &     \textbf{99.1} &     \textbf{56.6} &      63.0  &  \textbf{94.6} \\
    \bottomrule
    \end{tabular}
    \caption{Top-1 classification accuracy (\%) for event stream classification tasks.}
    \label{tab:event-results}
\end{table}

\bibliographystyle{splncs}
\bibliography{cdc.bib}

\newpage
\section{Supplementary Material}

\subsection{Summary of Notation}
\begin{table}[ht!]
    \centering
    \begin{tabular}{ll}
        \toprule
        \multicolumn{2}{l}{\textbf{Dimensions}} \\
        \hline
        $D$     & Physical dimensionality of point cloud \\
        $Q$     & Number of input channels \\
        $P$     & Number of output channels \\
        $\Sj$   & Size of input cloud \\
        $\Si$   & Size of output cloud \\
        $E$     & Number of edges \\
        $M$     & Number of basis functions \\ 
        \hline
        \multicolumn{2}{l}{\textbf{Sets}} \\
        \hline
        $\Xj \subset \mathbb{R}^{D}$  & Input cloud coordinates \\
        $\Xi \subset \mathbb{R}^{D}$  & Output cloud coordinates \\
        $\mathcal{N}_i \subseteq \Xj$  & Set of inputs in neighborhood of $\xi$ \\
        \hline
        \multicolumn{2}{l}{\textbf{Tensors}} \\
        \hline
        $\xj \in \Xj$ & $j^{\text{th}}$ input coordinate \\
        $\xi \in \Xi$ & $i^{\text{th}}$ output coordinate \\
        $\Delta x_{ij} \in \mathbb{R}^D$ & Edge vector: $\xi - \xj, \, \xj \in \mathcal{N}_i$ \\
        $\fjj \in \mathbb{R}$     & Input feature associated with $\xj$ \\
        $\fii \in \mathbb{R}$     & Output feature associates with $\xi$ \\
        $f \in \mathbb{R}^{\Sj}$    & Single-channel input feature for input cloud $\Xj$ \\
        $f^\prime \in \mathbb{R}^{\Si}$ & Single-channel output feature for output clouse $\Xi$ \\
        $\Fjj \in  \mathbb{R}^{\Sj \times Q}$ & Multi-channel input features \\
        $\Fii \in \mathbb{R}^{\Si \times P}$ & Multi-channel output feature \\
        $\Thetam \in \mathbb{R}^{Q \times P}$ & kernel parameters associated with $m^{\text{th}}$ basis fn \\
        $\Nm \in \mathbb{R}^{\Si \times \Sj}$ & Neighborhood matrix \\  
        \bottomrule
    \end{tabular}
    \caption{Summary of notation.}
    \label{tab:notation}
\end{table}
\newpage 
\subsection{Additional Point Cloud Network Details}
Pseudo-code for Iterative Farthest Point (IFP) variants and rejection sampling are given in Algorithms \ref{alg:ifp} through ref{alg:approx-ifp-rej}.

\begin{figure}
    \centering
    \setlength\tabcolsep{-2pt}
    \begin{tabular}{cc}
        Random & Rejection \\
        \includegraphics[width=0.48\textwidth]{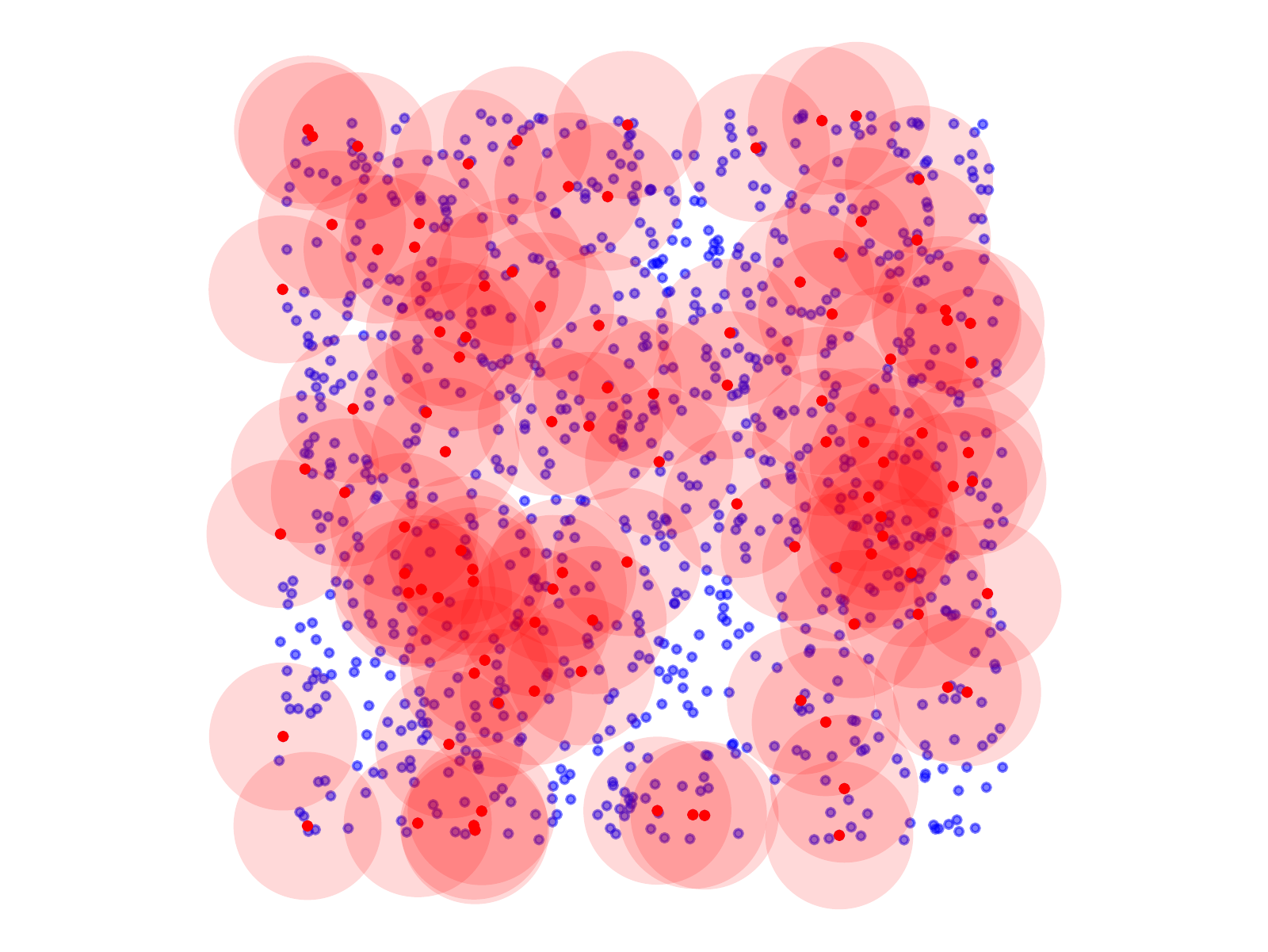} &
        \includegraphics[width=0.48\textwidth]{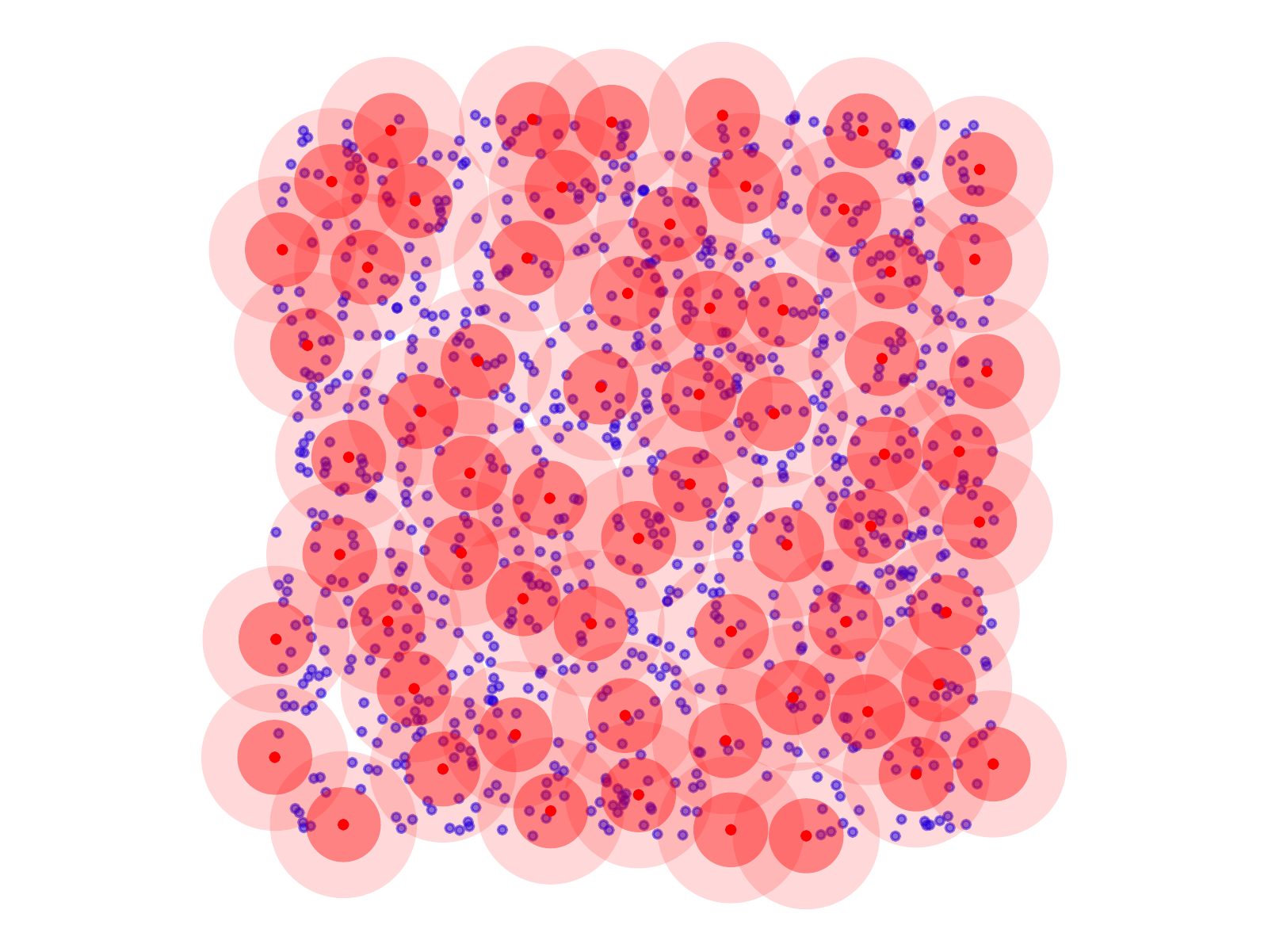} \\
        \includegraphics[width=0.48\textwidth]{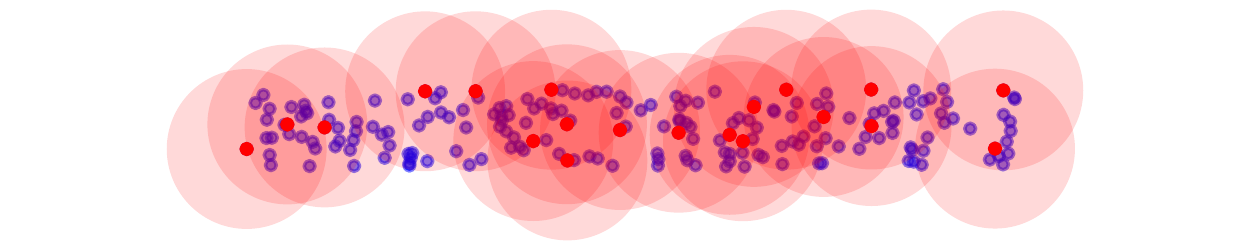} &
        \includegraphics[width=0.48\textwidth]{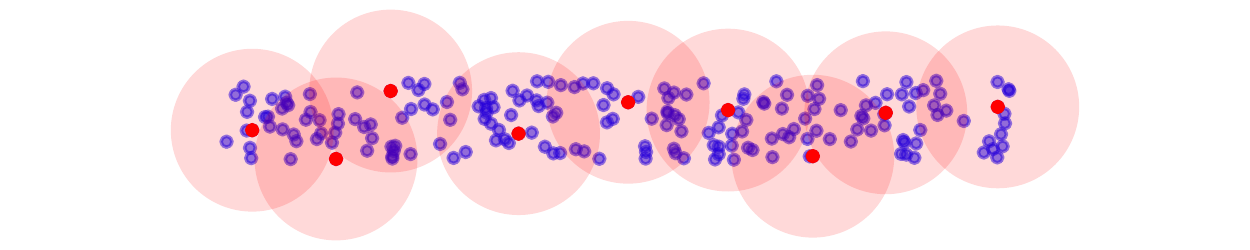}
    \end{tabular}
    \setlength\tabcolsep{6pt}
    \caption{Output cloud (red dots) resulting from different sampling schemes applied to input clouds (blue) and the corresponding neighborhoods (light red circles). From the top left image, we can see random sampling can result in some regions being under-sampled. This is particularly problematic for networks with subsequent up-sampling, where some blue points have no red points in their own neighborhood. The number of sampled points is not fixed for rejection sampling, so significantly less points will be sampled from pointy surfaces (bottom). By construction, none of the dark red circles (top right, half base radius) overlap, so the total number of possible sample points is limited by ball packing theorems.}
    \label{fig:occupancy}
\end{figure}

Select differences between rejection sampling and random sampling are given in Figure \ref{fig:occupancy}.

A diagram of our large point cloud network is given in Figure \ref{fig:point-cloud-network}.

\newpage

\begin{minipage}[t]{0.48\textwidth}
    \vspace{0pt}
    \begin{algorithm}[H]
        \SetAlgoLined
        \SetKwInOut{Inputs}{Inputs}
        \Inputs{
            $\Xj$ input point cloud\\
            $\Si$ output size}
        \KwResult{$\Xi$: sampled points}
         $\Xi \gets []$\;
         $\Sj \gets \text{size}(\Xj)$\;
         $d_{\text{min}} \gets \infty \times \text{ones}(\Sj)$\;
         \For{$i$ in $\text{range}(\Si)$}{
          $j \gets \text{argmin}(d_{\text{min}})$\;
          $\Xi.\text{append}(\xj)$\;
          $d_{\text{min}} \gets \min(d_{min}, d(\Xj, \xj))$\;
         }
         \caption{IFP}
         \label{alg:ifp}
    \end{algorithm}
    \begin{algorithm}[H]
        \SetAlgoLined
        \SetKwInOut{Inputs}{Inputs}
        \Inputs{
            $\Xj$ input point cloud\\
            $\Si$ output size \\
            $\mathcal{N}(\cdot)$ neighborhood fn\\
            $Q$ priority queue}
        \KwResult{$\Xi$: sampled points}
         $\Xi \gets []$\;
         \For{$j$ in $\text{range}(\Si)$}{
            $\xi \gets Q$.pop()\;
            $\Xi.\text{append}(\xi)$\;
            \For{$x_n$ in $\mathcal{N}(\xi)$}{
                $Q$.update($x_n$, $d(x_n, \xi)$)
            }
         }
         \caption{Approx. IFP}
         \label{alg:approx-ifp}
    \end{algorithm}
    \begin{algorithm}[H]
        \SetAlgoLined
        \SetKwInOut{Inputs}{Inputs}
        \Inputs{
            $\Xj$ input point cloud\\
            $\Si$ output size\\
            $\mathcal{N}(\cdot)$ neighborhood fn}
        \KwResult{$\Xi$: sampled points}
         $\Sj \gets \text{size}(\Xj)$\;
         $Q \gets \text{Priority Queue}(\infty \times \text{ones}(\Sj), \Xj)$\;
         $\Xi \gets \text{Approx. IFP}(\Xj, \Si, \mathcal{N}, Q)$
         \caption{Approx. IFP (without rej.)}
         \label{alg:approx-ifp-no-rej}
    \end{algorithm}
\end{minipage}
\hfill
\begin{minipage}[t]{0.48\textwidth}
    \vspace{0pt}
    \begin{algorithm}[H]
        \SetAlgoLined
        \SetKwInOut{Inputs}{Inputs}
        \Inputs{$\Xj$ input point cloud\\
                $\mathcal{N}(\cdot)$ neighborhood fn}
        \KwResult{\\
        \quad $\Xi$: sampled points \\
        \quad $d_{min}$: distance from each input \\
        \quad \quad point to closest output point}%
         $\Xi \gets []$\;
         $\Sj \gets \text{size}(\Xj)$\;
         $d_{\text{min}} \gets \infty \times \text{ones}(\Sj)$\;
         visited $\gets \text{False} \times \text{ones}(\Sj)$\;
         \For{$\xi$ in $\Xj$}{
          \If{$\text{visited}[i]$}{
            continue\;
          }
          $\Xi.\text{append}(\xi)$\;
          $\mathcal{N}_i \gets \mathcal{N}(\xi)$\;
          \For{$\xj$ in $\mathcal{N}_i$}{
            $\text{visited}[j] \gets \text{True}$\;
            $d_{\text{min}}[j] \gets \min(d_{min}[j], d(\xi, \xj))$\;
          }%
         }%
         \caption{Rejection Sampling}
         \label{alg:rejection}
    \end{algorithm}
    \begin{algorithm}[H]
        \SetAlgoLined
        \SetKwInOut{Inputs}{Inputs}
        \Inputs{
            $\Xj$ input point cloud\\
            $\Si$ output size\\
            $\mathcal{N}(\cdot)$ neighborhood fn}
        \KwResult{$\Xi$: sampled points}
         $\Xi_0, d_\text{min} \gets \text{Rejection Sampling}(\Xj, \Si, \mathcal{N})$\;
         $Q \gets \text{Priority Queue}(d_{\text{min}}, \Xj)$\;
         $\Si_1 \gets \Si - \text{size}(\Xi_0)$\;
         $\Xi_1 \gets \text{Approx. IFP}(\Xj, \Si_1, \mathcal{N}, Q)$\;
         $\Xi \gets \text{concatenate}(\Xi_0, \Xi_1)$\;
         \caption{Approx. IFP (with rej.)}
         \label{alg:approx-ifp-rej}
    \end{algorithm}
\end{minipage}%
\newpage
\begin{figure}[ht!]
    \centering
    \begin{subfigure}{.45\textwidth}
        \centering
        \resizebox{\columnwidth}{!}{
        \input{tikz/resnet-ball-identity.tikz}
        }
        \caption{In-place residual block}
    \end{subfigure}
    \hfill
    \begin{subfigure}{.45\textwidth}
        \centering
        \resizebox{\columnwidth}{!}{
        \input{tikz/resnet-ball-conv.tikz}
        }
        \caption{Down-sample residual block}
    \end{subfigure}

    \begin{subfigure}{\textwidth}
        \centering
        \resizebox{0.5\columnwidth}{!}{
        \input{tikz/ball-network.tikz}
        }
        \caption{Large Point Cloud Network, $r_0=0.1125$. Numbers in brackets represent output example dimensions. Dimensions with question marks (?) correspond to approximate number of points when using no point dropout. Dashed line corresponds to preprocessing/batching divide. BN is batch normalization, and Dropout uses a rate of 0.5}
    \end{subfigure}
    \caption{}
    \label{fig:point-cloud-network}
\end{figure}

\subsection{Additional Event Stream Network Details}
The Leaky Integrate and Fire (LIF) algorithm we used is given in Algorithm \ref{alg:lif}.

\begin{algorithm}[ht!]
    \SetAlgoLined
    \SetKwInOut{Inputs}{Inputs}
    \Inputs{
        $X$ input grid shape \\
        $t$ times for input events, sorted ascending\\
        $x$ coordinates for input events, same order as $t$\\
        $\mathcal{N}(\cdot)$ spatial neighborhood
        fn giving coordinates of receptive field \\
        $\tilde{t}$ decay time \\
        $v_\text{thresh}$ spike threshold \\
        $v_\text{reset}$ reset potential}
    \KwResult{$t_{\text{out}}, x_{\text{out}}$: time and coordinates of output stream.}
     $x_{\text{out}} \gets []$\;
     $t_{\text{out}} \gets []$\;
     $V \gets \text{zeros}(X)$\;
     $T \gets \text{zeros}(X)$\;
     $\Sj \gets \text{size}(x)$\;
     \For{$i$ in $\text{range}(\Sj)$}{
      $t_i \gets t[i]$\;
      $x_i \gets x[i]$\;
      $n \gets \text{size}(\mathcal{N}(x_i))$\;
      \For{$x_j$ in $\mathcal{N}(x_i)$}{
      $v \gets V[x_j] \exp\left(-(t_i - T[x_j]) / \tilde{t}\right) + \frac{1}{n}$\;
          \If{$v > v_{\text{thresh}}$}{
            $v \gets v_{\text{reset}}$\;
            $t_{\text{out}}.\text{append}(t_i)$\;
            $x_{\text{out}}.\text{append}(x_j)$\;
          }
          $V[x_j] \gets v$\;
          $T[x_j] \gets t_i$\;
        }
     }
     \caption{Leaky Integrate and Fire (LIF)}
     \label{alg:lif}
\end{algorithm}

We down-sampled examples from the two highest-resolution datasets -- N-Caltech101 and ASL-DVS -- by a factor of 2 in each dimension. We performed basic data augmentation involving small rotations ($-22.5^\circ$ to $22.5^\circ$), time/polarity reversal for all datasets except ASL-DVS and left-right flips for CIFAR-10-DVS and N-Caltech101. No data augmentation was applied to ASL-DVS. We computed neighborhood information for N-MNIST online and offline with 8 augmented repeats for MNIST-DVS, CIFAR10-DVS and NCaltech101-DVS.

For the small number of examples with more than 300,000 events we took the first 300,000. Apart from this infrequent cropping, we use all events in all examples.

All models were trained with Adam optimizer, initial learning rate $1e-3$, $\beta_1=0.9$, $\beta_2=0.999$, $\epsilon=1e-7$. We trained our ASL-DVS model for 100 epochs with a fixed learning rate. For all others, we decay the learning rate by a factor of 5 after the training accuracy fails to increase for 10 epochs, and run until learning ceases as a result of several such decays.

Dataset summary statistics and select model hyper-parameters parameters given in Table \ref{tab:event-data-summary}.

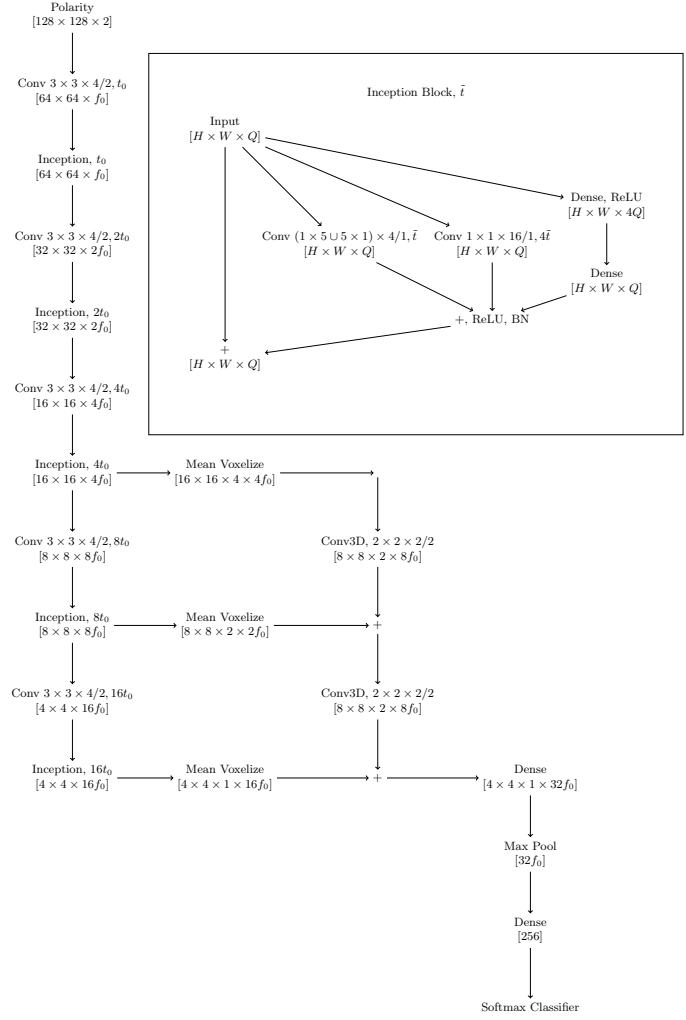
\begin{figure}[ht!]
    \centering
    \resizebox{0.75\columnwidth}{!}{
        \input{tikz/stream-network.tikz}
    }
    \caption{Network architecture for event stream inference for CIFAR10-DVS. Conv $h \times w \times t / S, \tilde{t}$ is a down-sampling convolution with spatial stride $S$, spatial kernel shape $h \times w$ and temporal kernel size $t$, \ie $M_u=hw$, $M_v=t$. The output stream is the result of LIF subsampling with the same spatial kernel size and decay time $\tilde{t}$. Edges with $\Delta t > 4\tilde{t}$ are cropped, and convolutions use $\Delta t$ scaled by $\tilde{t}$. Each down-sampling convolution, the pre-max-pooling dense layer and the final dense layer are all followed by ReLU, batch normalization and dropout with rate 0.5. Each mean voxelization is followd by batch normalization, and each Conv3D is followed by ReLU  and batch normalization.}
    \label{fig:stream-network}
\end{figure}

A diagram of the model used for CIFAR10-DVS is given in Figure \ref{fig:stream-network}.

\begin{table}[ht!]
    \centering
    \resizebox{\columnwidth}{!}{
    \begin{tabular}{l|ccccc}
    \toprule
        Dataset & N-MNIST & MNIST-DVS & CIFAR10-DVS & N-Caltech101 & ASL-DVS \\ 
        \hline
        \# Classes  & 10 & 10 & 10 & 101 & 24 \\
        Resolution & $34 \times 34$ & $128 \times 128$ & $128 \times 128$ & $174 \times 234$ & $180 \times 240$ \\
        \# Train examples & 60,000 &   9,000 &   9,000 & 7,838   &  80,640 \\
        Median \# events     &  4,196 &  70,613 & 203,301 & 104,904 &  17,078 \\
        Mean \# events       &  4,171 &  73,704 & 204,979 & 115,382 &  28,120 \\
        Max \# events        &  8,183 & 151,124 & 422,550 & 428,595 & 470,435 \\
        \hline
        Batch Size        &     32 &      32 &       16 &       8 &       8 \\
        Spike Threshold, $v_{\text{thresh}}$   &    1.5 &     1.5 &     1.6 &     1.25 &     1.0 \\
        Reset Potential, $v_{\text{reset}}$   &   -3.0 &    -2.0 &    -3.0 &    -2.0 &    -3.0 \\
        Initial Decay Time, $t_0$ &  2,000 &  10,000 &   4,000 &   1,000 &   1,000 \\
        Initial Filters, $f_0$   &     32 &      8 &        8 &      16 &      16 \\
        \# Down Samples   &      3 &      5 &        5 &       5 &       5 \\
        \hline
        \textbf{Data Augmentation} & & & & & \\
        Rotation up to $\pm22.5^\circ$ & Yes & Yes & Yes & Yes & No \\
        Flip left-right & No & No & Yes & Yes & No \\
        Flip time/polarity & Yes & Yes & Yes & Yes & No \\
        Preprocessing repeats  & $\infty$ (online) & 8 & 8 & 8 & 1 \\
    \bottomrule
    \end{tabular}
    }
    \vspace*{1mm}
    \caption{Event stream dataset summary statistics and model/data augmentation hyperparameters.}
    \label{tab:event-data-summary}
\end{table}

\end{document}

%% file: tikz/conceptual.tikz
\begin{tikzpicture}[{every node/.style}={scale=0.5}, scale=0.5]
	\begin{pgfonlayer}{nodelayer}
		\node [] (x0) at (0, -0) {$\mathcal{X}_0$};
		\node [] (x1) at (0, -4) {$\mathcal{X}_1$};
		\node [] (x2) at (0, -8) {$\mathcal{X}_2$};
		\node [] (s0) at (0, -2) {sample};
		\node [] (s1) at (0, -6) {sample};
		\node [] (c00) at (4, -2) {Featureless Conv.};
		\node [] (c11) at (4, -6) {In-Place Conv.};
		\node [] (c01) at (4, -4) {Down-Sample Conv.};
		\node [] (c12) at (4, -8) {Down-Sample Conv.};
		\node [] (f01) at (4, -3) {$F_{01}$};
		\node [] (f11) at (4, -5) {$F_{11}$};
		\node [] (f12) at (4, -7) {$F_{12}$};
		\node [] (f22) at (4, -9) {$\vdots$};
		\node [] (t00) at (8, -2) {$\Phi_0$};
		\node [] (t01) at (8, -4) {$\Theta_{01}$};
		\node [] (t11) at (8, -6) {$\Theta_{11}$};
		\node [] (t12) at (8, -8) {$\Theta_{12}$};
	\end{pgfonlayer}
	\begin{pgfonlayer}{edgelayer}
		\draw[->] (x0) -> (s0);
		\draw[->] (x1) -> (s1);

		\draw[->] (s0) -> (x1);
		\draw[->] (s1) -> (x2);

		\draw[->] (x0) -> (c00);
		\draw[->] (x1) -> (c11);
		\draw[->] (x0) -> (c01);
		\draw[->] (x1) -> (c12);

		\draw[->] (t00) -> (c00);
		\draw[->] (t01) -> (c01);
		\draw[->] (t11) -> (c11);
		\draw[->] (t12) -> (c12);

		\draw[->] (c00) -> (f01);
		\draw[->] (f01) -> (c01);
		\draw[->] (c01) -> (f11);
		\draw[->] (f11) -> (c11);
		\draw[->] (c11) -> (f12);
		\draw[->] (f12) -> (c12);
		\draw[->] (c12) -> (f22);

		\draw[->] (x1) -> (c01);
		\draw[->] (x2) -> (c12);

	\end{pgfonlayer}
\end{tikzpicture}

%% file: tikz/meta.tikz
\begin{tikzpicture}[{every node/.style}={scale=0.5}, scale=0.5]
	\begin{pgfonlayer}{nodelayer}
		\node [] (ds) at (0, 0) {Dataset};
		\node [] (aug) at (0, -2) {Augmentation};
		\node [] (x0) at (0, -4) {$\mathcal{X}_{0}$};
		\node [] (x1) at (0, -8) {$\mathcal{X}_{1}$};
		\node [] (x2) at (0, -12) {$\vdots$};
		\node [] (kd0) at (0, -6) {KDTree Search};
		\node [] (kd1) at (0, -10) {KDTree Search};
		\node [] (n00) at (4, -6) {$N_{00}$};
		\node [] (n01) at (4, -8) {$N_{01}$};
		\node [] (n11) at (4, -10) {$N_{11}$};
		\node [] (n12) at (4, -12) {$N_{12}$};
		\node [style=none] (batchlabel) at (5.75, -1.5) {Batching};
		\node [style=none] (batch0) at (5.75, -2) {};
		\node [style=none] (batch1) at (5.75, -14) {};
		\node [] (bd00) at (8, -6) {Block Diag.};
		\node [] (bd01) at (8, -8) {Block Diag.};
		\node [] (bd11) at (8, -10) {Block Diag.};
		\node [] (bd12) at (8, -12) {Block Diag.};
		\node [style=none] (c00) at (13, -6) {Featureless Conv.};
		\node [style=none] (c01) at (13, -8) {Down-Sample Conv.};
		\node [style=none] (c11) at (13, -10) {In-Place Conv.};
		\node [style=none] (c12) at (13, -12) {Down-Sample Conv.};
		\node [] (t00) at (16, -6) {$\Phi_{0}$};
		\node [] (t01) at (16, -8) {$\Theta_{01}$};
		\node [] (t11) at (16, -10) {$\Theta_{11}$};
		\node [] (t12) at (16, -12) {$\Theta_{12}$};
		\node [] (f01) at (13, -7) {$F_{01}$};
		\node [] (f11) at (13, -9) {$F_{11}$};
		\node [] (f12) at (13, -11) {$F_{12}$};
		\node [] (f22) at (13, -13) {$\vdots$};
	\end{pgfonlayer}
	\begin{pgfonlayer}{edgelayer}
		\draw[dashed] (-2, -1) rectangle (5, -14);
		\node [] (prebatch) at (1.5, -14.5) {Pre-batch (CPU)};
		\draw[dashed] (6.5, -5) rectangle (9.5, -14);
		\node [] (prebatch) at (8, -14.5) {Post-batch (CPU)};
		\draw[dashed] (10.5, -5) rectangle (16.5, -14);
		\node [] (prebatch) at (13, -14.5) {Learned (Accelerator)};
		\draw[->] (ds) -- (aug);
		\draw[->] (aug) -- (x0);
		\draw[->] (x0) -- (kd0);
		\draw[->] (x1) -- (kd1);
		\draw[->] (kd0) -- (x1);
		\draw[->] (kd1) -- (x2);
		\draw[->] (kd0) -- (n00);
		\draw[->] (kd0) -- (n01);
		\draw[->] (kd1) -- (n11);
		\draw[->] (kd1) -- (n12);
		\draw[->] (n00) -- (bd00);
		\draw[->] (n01) -- (bd01);
		\draw[->] (n11) -- (bd11);
		\draw[->] (n12) -- (bd12);

		\draw[->] (bd00) -- (c00);
		\draw[->] (t00) -- (c00);
		\draw[->] (c00) -- (f01);

		\draw[->] (f01) -- (c01);
		\draw[->] (bd01) -- (c01);
		\draw[->] (t01) -- (c01);
		\draw[->] (c01) -- (f11);

		\draw[->] (f11) -- (c11);
		\draw[->] (bd11) -- (c11);
		\draw[->] (t11) -- (c11);
		\draw[->] (c11) -- (f12);

		\draw[->] (f12) -- (c12);
		\draw[->] (bd12) -- (c12);
		\draw[->] (t12) -- (c12);
		\draw[->] (c12) -- (f22);

		\draw[dotted] (batch0) -- (batch1);
	\end{pgfonlayer}
\end{tikzpicture}

%% file: tikz/resnet-ball-identity.tikz
\begin{tikzpicture}[every text node part/.style={align=center}]
	\begin{pgfonlayer}{nodelayer}
		\node [draw] (0) at (0, 7) {$\Sj \times Q$};
		\node [] (2) at (2, 6.5) {Dense\\ReLU\\BN};
		\node [draw] (3) at (4, 6) {$\Sj \times Q / 4$};
		\node [] (4) at (4, 5) {Conv\\ReLU\\BN};
		\node [draw] (5) at (4, 4) {$\Sj \times Q / 4$};
		\node [] (6) at (4, 3) {Dense\\BN\\Dropout};
		\node [draw] (7) at (4, 2) {$\Sj \times Q$};
		\node [] (8) at (0, 1) {$+$\\ReLU};
		\node [draw] (12) at (0, -0) {$\Sj \times Q$};
	\end{pgfonlayer}
	\begin{pgfonlayer}{edgelayer}
		\draw [->] (0) to (2);
		\draw [->] (2) to (3);
		\draw [->] (3) to (4);
		\draw [->] (4) to (5);
		\draw [->] (5) to (6);
		\draw [->] (6) to (7);
		\draw [->] (7) to (8);
		\draw [->] (0) to (8);
		\draw [->] (8) to (12);
	\end{pgfonlayer}
\end{tikzpicture}

%% file: tikz/resnet-ball-conv.tikz
\begin{tikzpicture}[every text node part/.style={align=center}]
	\begin{pgfonlayer}{nodelayer}
		\node [draw] (0) at (0, 7) {$\Sj \times Q$};
		\node [] (1) at (0, 5) {Sample};
		\node [] (2) at (2, 6.5) {Dense\\ReLU\\BN};
		\node [draw] (3) at (4, 6) {$\Sj \times P / 4$};
		\node [] (4) at (4, 5) {Conv\\ReLU\\BN};
		\node [draw] (5) at (4, 4) {$\Si \times P / 4$};
		\node [] (6) at (4, 3) {Dense\\BN\\Dropout};
		\node [draw] (7) at (4, 2) {$\Si \times P$};
		\node [] (8) at (0, 1) {$+$\\ReLU};
		\node [draw] (9) at (0, 4) {$\Si \times Q$};
		\node [] (10) at (0, 3) {Dense\\BN\\Dropout};
		\node [draw] (11) at (0, 2) {$\Si \times P$};
		\node [draw] (12) at (0, -0) {$\Si \times P$};
	\end{pgfonlayer}
	\begin{pgfonlayer}{edgelayer}
		\draw [->] (0) to (2);
		\draw [->] (2) to (3);
		\draw [->] (3) to (4);
		\draw [->] (4) to (5);
		\draw [->] (5) to (6);
		\draw [->] (6) to (7);
		\draw [->] (7) to (8);
		\draw [->] (0) to (1);
		\draw [->] (1) to (9);
		\draw [->] (9) to (10);
		\draw [->] (10) to (11);
		\draw [->] (11) to (8);
		\draw [->] (8) to (12);
	\end{pgfonlayer}
\end{tikzpicture}

%% file: tikz/ball-network.tikz
\begin{tikzpicture}[every text node part/.style={align=center}]
	\begin{pgfonlayer}{nodelayer}
		\node [draw] (0) at (0, -0) {$\mathcal{X}_0$ \\ $[1024? \times 3]$};
		\node [draw] (1) at (0, -4) {$\mathcal{X}_1$ \\ $[256? \times 3]$};
		\node [draw] (2) at (0, -8) {$\mathcal{X}_2$ \\ $[64? \times 3]$};
		\node [draw] (3) at (0, -12) {$\mathcal{X}_3$ \\ $[16? \times 3]$};
		\node [draw] (4) at (0, -2) {Sample};
		\node [draw] (5) at (0, -6) {Sample};
		\node [draw] (6) at (0, -10) {Sample};
		\node [draw] (7) at (2, -1) {Ball \\$r = r_0$};
		\node [draw] (8) at (2, -5) {Ball \\$r = 2r_0$};
		\node [draw] (9) at (2, -9) {Ball \\$r = 4r_0$};
		\node [draw] (10) at (2, -13) {Ball \\$r = 8r_0$};
		\node [] (11) at (3.5, -4) {};
		\node [] (12) at (3.25, -8) {};
		\node [] (13) at (3.25, -12) {};
		\node [] (14) at (-2, -0) {};
		\node [] (15) at (-2, -3.375) {};
		\node [] (16) at (3.5, -3.375) {};
		\node [] (17) at (-2, -4) {};
		\node [] (18) at (-2, -7.375) {};
		\node [] (19) at (3.25, -7.375) {};
		\node [] (20) at (-2, -8) {};
		\node [] (21) at (-2, -11.375) {};
		\node [] (22) at (3.25, -11.375) {};
		\node [draw] (23) at (8, -3.75) {$N_{01}$ \\ $[10 \times 256? \times 1024?]$};
		\node [draw] (24) at (8, -5) {$N_{11}$ \\ $[10 \times 256? \times 256?]$};
		\node [draw] (25) at (8, -7.75) {$N_{12}$ \\ $[10 \times 64? \times 256?]$};
		\node [draw] (26) at (8, -9) {$N_{22}$ \\ $[10 \times 64? \times 64?]$};
		\node [draw] (27) at (8, -11.75) {$N_{23}$ \\ $[10 \times 16? \times 64?]$};
		\node [draw] (28) at (8, -13) {$N_{33}$ \\ $[10 \times 16? \times 16?]$};
		\node [draw] (29) at (12, -3.75) {Res Block \\ $[256? \times 64]$};
		\node [draw] (30) at (12, -5) {Res Block \\ $[256? \times 64]$};
		\node [draw] (31) at (12, -7.75) {Res Block \\ $[64? \times 128]$};
		\node [draw] (32) at (12, -9) {Res Block \\ $[64? \times 128]$};
		\node [draw] (33) at (12, -11.75) {Res Block \\ $[16? \times 256$]};
		\node [draw] (34) at (12, -13) {Res Block \\ $[16? \times 256]$};
		\node [draw] (35) at (12, -15) {Dense \\ $[16? \times 1024]$};
		\node [] (36) at (9.75, 1) {};
		\node [] (37) at (9.75, -21) {};
		\node [draw] (38) at (4.25, -3.75) {Ball \\$r = \sqrt{2}r_0$};
		\node [draw] (39) at (4.25, -7.75) {Ball \\$r = 2\sqrt{2}r_0$};
		\node [draw] (40) at (4.25, -11.75) {Ball \\$r = 4\sqrt{2}r_0$};
		\node [draw] (41) at (12, -16.5) {Global Pooling \\ $[1024]$};
		\node [draw] (42) at (12, -18) {ReLU, BN, Dropout, Dense \\ $[256]$};
		\node [draw] (43) at (12, -20) {ReLU, BN, Dropout, Dense \\ $[40]$};
		
		\node [draw] (50) at (8, -1) {$N_{00}$ \\ $[10 \times 1024? \times 1024?]$};
		
		\node [draw] (51) at (12, -1) {Feat.less Conv \\ $[1024 \times 32]$};
	\end{pgfonlayer}
	\begin{pgfonlayer}{edgelayer}
		\draw [dashed] (36) to (37);
		\draw [->] (0) to (7);
		\draw [->] (7) to (4);
		\draw [->] (0) to (4);
		\draw [->] (4) to (1);
		\draw [->] (1) to (11);
		\draw [->] (1) to (8);
		\draw [->] (8) to (5);
		\draw [->] (1) to (5);
		\draw [->] (5) to (2);
		\draw [->] (2) to (12);
		\draw [->] (2) to (9);
		\draw [->] (9) to (6);
		\draw [->] (2) to (6);
		\draw [->] (6) to (3);
		\draw [->] (3) to (13);
		\draw [->] (3) to (10);
		\draw [->] (38) to (23);
		\draw [->] (39) to (25);
		\draw [->] (40) to (27);
		\draw [->] (8) to (24);
		\draw [->] (9) to (26);
		\draw [->] (10) to (28);
		\draw [->] (0) to (14.center) to (15.center) to (16);
		\draw [->] (1) to (17.center) to (18.center) to (19);
		\draw [->] (2) to (20.center) to (21.center) to (22);
		\draw [->] (23) to (29);
		\draw [->] (24) to (30);
		\draw [->] (25) to (31);
		\draw [->] (26) to (32);
		\draw [->] (27) to (33);
		\draw [->] (28) to (34);
		\draw [->] (29) to (30);
		\draw [->] (30) to (31);
		\draw [->] (31) to (32);
		\draw [->] (32) to (33);
		\draw [->] (33) to (34);
		\draw [->] (34) to (35);
		\draw [->] (35) to (41);
		\draw [->] (41) to (42);
		\draw [->] (42) to (43);
		
		\draw [->] (7) to (50);
		\draw [->] (50) to (51);
		\draw [->] (51) to (29);
	\end{pgfonlayer}
\end{tikzpicture}

%% file: tikz/stream-network.tikz
\begin{tikzpicture}[every text node part/.style={align=center}]
	\begin{pgfonlayer}{nodelayer}
		\node [] (0) at (0, -0) {Polarity\\$[128 \times 128 \times 2]$};
		\node [] (1) at (0, -2) {Conv $3 \times 3 \times 4/ 2, t_0$\\$[64 \times 64 \times f_0]$};
		\node [] (2) at (0, -6) {Conv $3 \times 3 \times 4/ 2, 2t_0$\\$[32 \times 32 \times 2f_0]$};
		\node [] (3) at (0, -10) {Conv $3 \times 3 \times 4/ 2, 4t_0$\\$[16 \times 16 \times 4f_0]$};
		\node [] (4) at (0, -14) {Conv $3 \times 3 \times 4/ 2, 8t_0$\\$[8 \times 8 \times 8f_0]$};
		\node [] (5) at (0, -18) {Conv $3 \times 3 \times 4/ 2, 16t_0$\\$[4 \times 4 \times 16f_0]$};
		\node [] (6) at (0, -4) {Inception, $t_0$\\$[64 \times 64 \times f_0]$};
		\node [] (7) at (0, -8) {Inception, $2t_0$\\$[32 \times 32 \times 2f_0]$};
		\node [] (8) at (0, -12) {Inception, $4t_0$\\$[16 \times 16 \times 4f_0]$};
		\node [] (9) at (0, -16) {Inception, $8t_0$\\$[8 \times 8 \times 8f_0]$};
		\node [] (10) at (0, -20) {Inception, $16t_0$\\$[4 \times 4 \times 16f_0]$};
		\node [] (11) at (4, -12) {Mean Voxelize\\$[16 \times 16 \times 4 \times 4f_0]$};
		\node [] (12) at (4, -16) {Mean Voxelize\\$[8 \times 8 \times 2 \times 2f_0]$};
		\node [] (13) at (4, -20) {Mean Voxelize\\$[4 \times 4 \times 1 \times 16f_0]$};
		\node [] (14) at (8, -12) {};
		\node [] (15) at (8, -16) {+};
		\node [] (16) at (8, -20) {+};
		\node [] (17) at (8, -14) {Conv3D, $2\times 2\times 2 / 2$\\$[8\times 8 \times 2 \times 8f_0]$};
		\node [] (18) at (8, -18) {Conv3D, $2\times 2\times 2 / 2$\\$[8\times 8 \times 2 \times 8f_0]$};
		\node [] (19) at (12, -20) {Dense\\$[4 \times 4 \times 1 \times 32f_0]$};
		\node [] (20) at (12, -22) {Max Pool\\$[32f_0]$};
		\node [] (21) at (12, -24) {Dense\\$[256]$};
		\node [] (22) at (12, -26) {Softmax Classifier};
		\node [] (27) at (4, -3) {Input\\$[H \times W \times Q]$};
		\node [] (28) at (4, -9) {+\\$[H \times W \times Q]$};
		\node [] (29) at (7, -6) {Conv $(1 \times 5 \cup 5 \times 1) \times 4 / 1,\tilde{t}$ \\$[H \times W \times Q]$};
		\node [] (30) at (11, -6) {Conv $1 \times 1 \times 16 / 1,4\tilde{t}$\\$[H \times W \times Q]$};
		\node [] (31) at (14, -5) {Dense, ReLU\\$[H \times W \times 4Q]$};
		\node [] (32) at (14, -7) {Dense\\$[H \times W \times Q]$};
		\node [] (33) at (11, -8) {+, ReLU, BN};
		\node [] (40) at (9, -2) {Inception Block, $\tilde{t}$};
	\end{pgfonlayer}
	\begin{pgfonlayer}{edgelayer}
		\draw [->] (0) to (1);
		\draw [->] (1) to (6);
		\draw [->] (6) to (2);
		\draw [->] (2) to (7);
		\draw [->] (7) to (3);
		\draw [->] (3) to (8);
		\draw [->] (8) to (4);
		\draw [->] (4) to (9);
		\draw [->] (9) to (5);
		\draw [->] (5) to (10);
		\draw [->] (8) to (11);
		\draw [->] (9) to (12);
		\draw [->] (10) to (13);
		\draw [->] (11) to (14);
		\draw [->] (12) to (15);
		\draw [->] (13) to (16);
		\draw [->] (14) to (17);
		\draw [->] (17) to (15);
		\draw [->] (15) to (18);
		\draw [->] (18) to (16);
		\draw [->] (16) to (19);
		\draw [->] (19) to (20);
		\draw [->] (20) to (21);
		\draw [->] (21) to (22);
		\draw (2, -1) -- (2, -11) -- (16, -11) -- (16, -1) -- cycle;
		\draw [->] (27) to (29);
		\draw [->] (27) to (30);
		\draw [->] (27) to (31);
		\draw [->] (31) to (32);
		\draw [->] (29) to (33);
		\draw [->] (30) to (33);
		\draw [->] (32) to (33);
		\draw [->] (33) to (28);
		\draw [->] (27) to (28);
	\end{pgfonlayer}
\end{tikzpicture}